\documentclass{article}

\usepackage[final]{neurips_2024}

\usepackage[utf8]{inputenc} 
\usepackage[T1]{fontenc}    
\usepackage{hyperref}       
\usepackage{url}            
\usepackage{booktabs}       
\usepackage{amsmath,amssymb,amsfonts}       
\usepackage{nicefrac}       
\usepackage{microtype}      
\usepackage{xcolor}         
\usepackage{graphicx}
\usepackage{amsmath}
\usepackage{multirow}
\usepackage[linesnumbered]{algorithm2e}
\usepackage{textcomp}
\usepackage{float}
\usepackage{enumitem}
\usepackage{array}
\usepackage{setspace}
\usepackage[none]{hyphenat}
\usepackage{tabularx}

\author{%
Thi Thu Uyen Hoang  \quad Meenakshi Rajendran \quad Kun Zhang \quad Yuhan Wu \quad Viet Anh Nguyen \\
Saarland University \\
\texttt{\{thho00003, mera00002, kuzh00001, yuwu00001, ving00001\}@stud.uni-saarland.de} 
}

\begin{document}

\title{PDF Retrieval Augmented Question Answering}

\maketitle
\begin{abstract}
This paper presents an advancement in Question-Answering (QA) systems using a Retrieval Augmented Generation (RAG) framework to enhance information extraction from PDF files. Recognizing the richness and diversity of data within PDFs—including text, images, vector diagrams, graphs, and tables—poses unique challenges for existing QA systems primarily designed for textual content. We seek to develop a comprehensive RAG-based QA system that will effectively address complex multi-modal questions, where several data types are put together in the question. This is mainly achieved by refining approaches toward processing and integrating non-textual elements in PDFs into the RAG framework to derive precise and relevant answers, as well as finetuning large language models to adapt better into our system. We provide an in-depth experimental evaluation of our system, demonstrating its capabilities to extract accurate information that can be adapted to different types of content across PDFs. This work not only pushes the boundaries of retrieval-augmented QA systems but also lays a foundation for further research in multimodal data integration and processing. 

\end{abstract}

\section{Introduction}
Recent progress in machine learning and natural language processing has remarkably improved interactions with digital documents leading to better information retrieval systems. The most important aspect is the Retrieval Augmented Generation (RAG) framework \cite{a5} for QA systems, which combines both retrieval and generation-based approaches for handling difficult questions. In our work, we enhance the existing RAG-based QA system for information extraction through text, images, vector diagrams/graphs, and tables provided in PDFs.

RAG is designed to address the serious limitations of the large language models (LLMs) such as untruthfulness, false reasoning and hallucinations  \cite{a18}. RAG offers accurate and reliable solutions for generating contents and interacting with the users   \cite{a19}.
Retrieving information from PDF (Portable Document Format) has been drawing a huge attention in various academia and industries due to the data richness in PDF, from plain text, tables to high resolution images and intricate vector graphics, presenting an opportunity and a challenge at the same time. Traditional RAG-based QA systems focus primarily on text \cite{a7, a9, a12} while non-textual elements such as images, charts, tables and diagrams within PDFs are not thoroughly explored. Our objective is to address this gap by developing a comprehensive system capable of answering complex, multifaceted questions that necessitate the integration and interpretation of diverse data types.

To achieve this, we introduce an end-to-end system that retrieves and processes images, diagrams, graphs, and tables embedded within PDF documents, extending beyond the capabilities of conventional text-centric RAG models. We also implement preprocessing steps including the removal of headers and footers, conversion of PDFs to markdown for easier manipulation, image captioning and table reformatting to enhance data readability and retrieval accuracy. Finally, we fine-tune language models to be RAG-aware, ensuring a better understanding of our data format and document domain.

In the report, we discussed related work in \ref{sec:Related Work}, stated the objective of this project and implemention of preprocessing steps along with the model design in \ref{sec:PROCEDURE}. We present our experiment in \ref{sec:Experiments}, results in \ref{sec:Results} and conclude our report in \ref{sec:Conclusion}.

\section{Related Work}
\label{sec:Related Work}
The rapid advancement in multimodal QA stems from integrating RAG into multi-data modality frameworks. This section reviews relevant studies and developments, highlighting their contributions, methodologies and limitations of Integration of RAG with PDF Processing for QA.

\subsection{\textbf{Retrieval-Augmented Generation (RAG)}} Large language models (LLMs) have advanced AI but have limitations like hallucinations and inaccuracies. RAG improves text accuracy by leveraging retrieved documents. Corrective Retrieval Augmented Generation (CRAG) introduces evaluators to assess document quality and refine retrieval actions \cite{a16}. Unlike RAG, RAG-end2end \cite{a12} jointly trains retrievers and generators, enhancing open-domain question answering by updating all components, including external knowledge bases.

\subsection{\textbf{Question Answering (QA) with Language Models (LLMs)}}
 \cite{a15} democratizes advanced chat models, enhancing Llama's dialogue performance through fine-tuning and Self-Distillation with Feedback (SDF) further improves its capabilities. Comparing RAG and fine-tuning with synthetic data, fine-tuning shows significant performance improvements \cite{a13}. ChatQA \cite{a8} surpasses GPT-4 in retrieval-augmented generation and conversational QA. The Chain-of-Action (CoA) framework \cite{a20} addresses complex questions by decomposing them into reasoning chains, effectively tackling hallucinations. 

\subsection{\textbf{Multimodal Question Answering Systems}} Multimodal QA systems integrate diverse data modalities like text, tables, and images, improving real-world application accuracy. MMLLMs architecture \cite{a17} and the tool-interacting divide-and-conquer strategy \cite{a10} enhance reasoning and accuracy.

\subsection{\textbf{Integration of RAG with PDF Processing for QA}} PDFTriage \cite{a21} bridges this gap by enabling models to retrieve context based on both structure and content, but is challenged by the metadata variability, document format limitations, scalability, computational requirements, and datasets scope. A case study in the agricultural domain \cite{a1} demonstrated the approach combining RAG and Fine-Tuning exhibited superior performance when dealing with geographically specific knowledge. However, it does not fully leverage all available data types. This limitation reduces their effectiveness in scenarios that require integrated data sources, such as combining text with images and captions.

In our approach, we make use of the existing RAG model to answer queries relevant to PDF documents and overcome the above limitations regarding metadata handling, format compatibility and integration of different data sources from previous works.

\section{Methodology}
\label{sec:PROCEDURE}
\subsection{\textbf{Objective}}
A primary issue with RAG QA systems with PDFs is the automated extraction of elements like text, images and tables. Traditional text-based methods often fail to extract critical information resulting in less optimal performance \cite{a7}.  We aim to address the challenge of accurately extracting and processing different data types such as text, images, tables and diagrams. Our project objective is to develop a comprehensive system capable of answering complex, multifaceted questions that necessitate the integration, retrieval and interpretation of diverse data types.

\subsection{\textbf{System Design}}
\begin{figure}[h]
    \centering
    \includegraphics[width=\textwidth]{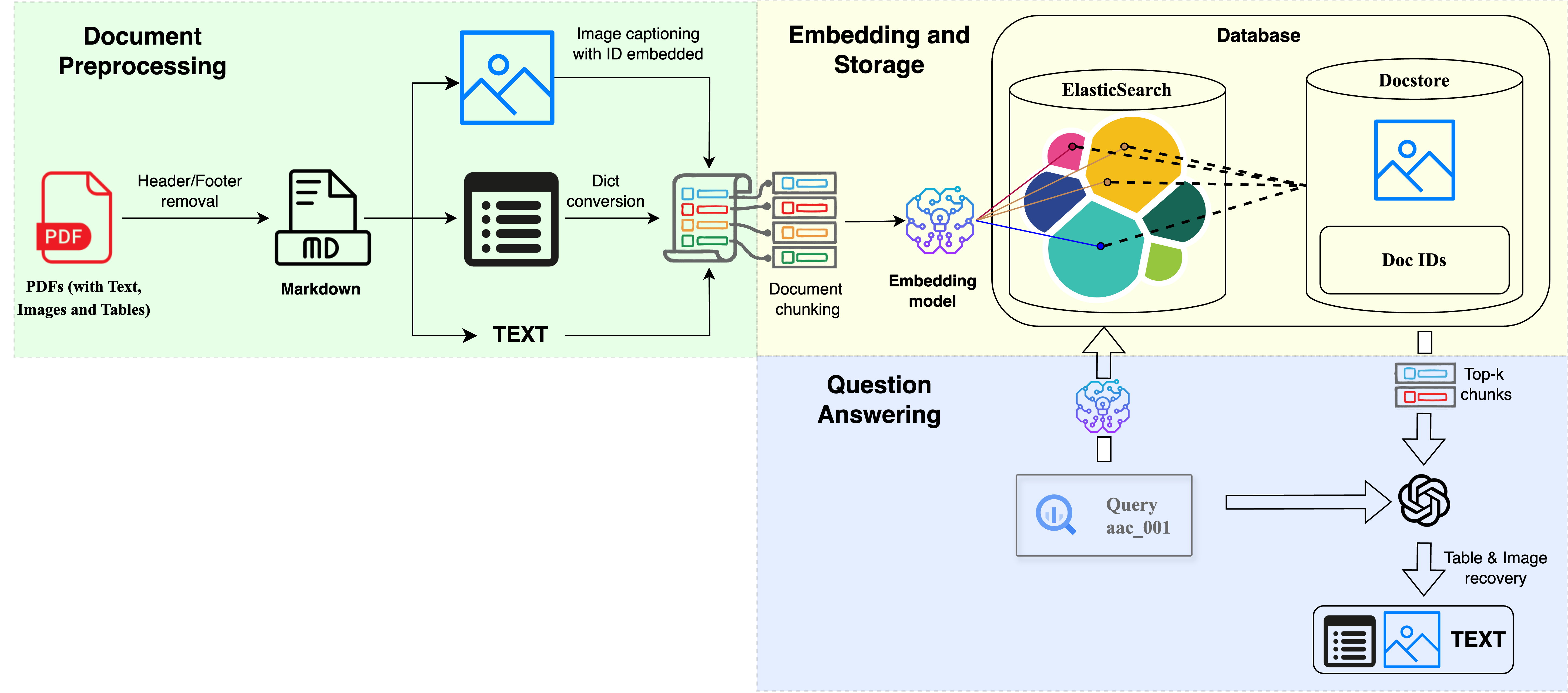}
    \caption{PIER-QA: PDF Integrated Enhanced Retrieval Question Answering}
    \label{fig:Figure_1 }
\end{figure}

Given a query $q$ and a set of PDF documents $\{D_1, D_2, \ldots, D_n\}$. The goal is to retrieve most accurate information from these documents and generate a precise answer $a$. We propose PDF Integrated Enhanced Retrieval Question Answering (PIER-QA) system consisting of three main components as shown in Figure \ref{fig:Figure_1 }.

\textbf{PDF Preprocessing.} Headers and footers are removed using DBSCAN clustering algorithm \cite{a2} which improves accuracy, ensuring documents are formatted for further processing. They are then converted to markdown through a machine-learning-based tool -- Marker \cite{a14}. Marker is a lightweight and easy to read format that simplifies further processing steps. In markdown format, we generate captions for images and compress markdown tables into a dictionary format for efficient storage and retrieval.

\textbf{Embedding and Storage.} The preprocessed markdown document is segmented into chunks of 1000 characters each, embedded by GTE-large \cite{a6} and stored using ElasticSearch \cite{a4} for efficient retrieval. RAPTOR \cite{a11} is used to enhance this process by indexing and clustering the chunks based on their semantics, improving the retrieval process. 

\textbf{Question answering.} Upon receiving a query from the user, ElasticEearch embeds the query, retrieves top-10 relevant chunks and uses the chunks as external knowledge to generate answers with our RAG-Aware LLM. If the LLM answer contains an image ID or a table, we recover the 
corresponding image and table format before displaying to the user. 

\subsubsection{\textbf{PDF Preprocessing}}
\begin{figure}[H]
    \centering
    \includegraphics[width=\textwidth]{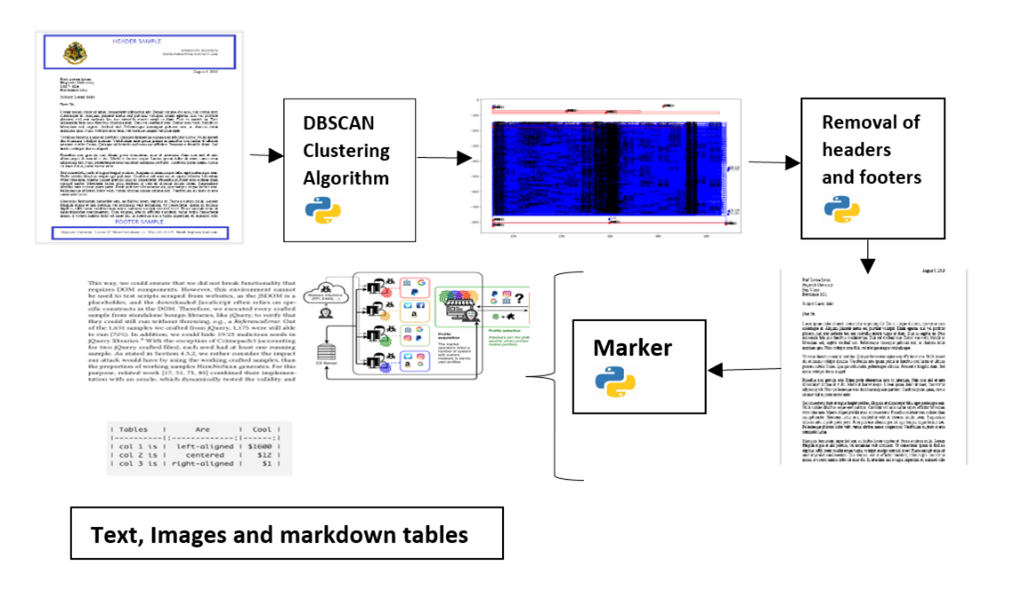} 
    \caption{PDF Preprocessing}
    \label{fig:figure_2}
\end{figure}

\textbf{Header and footer removal}

The removal of headers and footers is an important preprocessing step as they could interfere with the retrieval process by adding noise to the data which leads to less accurate results. Thus, by removing headers and footers, we obtain reliable cleaned pdf documents for further processing. We make an assumption that headers and footers coordinates are consistent across pages, i.e., at the top and bottom of the page, therefore by detecting this repeating pattern we will be able to remove headers and footers. As shown in Figure \ref{fig:figure_2}, we employ DBSCAN (Density-Based Spatial Clustering of Applications with Noise) which is well known for identifying areas of high density \cite{a2}. We use DBSCAN to cluster the bounding box of PDF elements, then remove the most frequent clusters across pages (marked by red boxes) while keeping the rest (blue boxes). We also notice the pattern can slightly varies, especially on long documents, therefore we apply the algorithm on each 10 pages instead of the entire document at once. Let $D_i$ be a PDF document. We denote the preprocessed version of $D_i$ as $\hat{D_i}$, the preprocessing involves:

\begin{equation}
    DBSCAN(D_i)\to\hat{D_i}^{clean}
\end{equation}

The implementation of DBSCAN clustering involves several key steps as shown in algorithm \ref{alg:one}. Initially, the PDF parsing library extracts the bounding boxes of all elements on each page, including text blocks, images, drawings, and other graphical elements. These bounding boxes are then used as input for the DBSCAN algorithm, which clusters them based on their spatial proximity. More details of DBSCAN hyperparameters can be found in Appendix~\ref{apdx:dbscan}.

\RestyleAlgo{ruled}
\begin{algorithm}
\caption{Header/Footer Removal Algorithm}\label{alg:one}
 Initialize DBSCAN parameters:\\
  minPts: minimum number of points to form a dense region\\
  eps: maximum distance between two points to be considered neighbors\\
Load PDF document\\
Extract text elements with spatial coordinates (x, y)\;
Cluster text elements using DBSCAN:\\
    clusters = DBSCAN(text\_elements, eps, minPts)\\
 \For{each cluster in clusters:}{
  Calculate cluster centroid\\
  \eIf{centroid is near the top or bottom of the page:}{
   Mark cluster as header or footer\\
   }{
   Mark cluster as main content\\
  }
  }
  \For{each page in PDF:}{
  Remove text elements marked as header or footer
  }
  Save the modified PDF document
\end{algorithm}

\textbf{PDF to Markdown conversion\\}
This step employs Marker (Venkatramana, 2023), a software utility tool designed for identifying and extracting various types of content from pdf documents helps in extracting and saving images along with markdown text and utilizes models wherever necessary to enhance speed and accuracy. We convert the cleaned document to markdown format including text  $T$, image $I$ , and tables $\tau$: 

\begin{equation}
    Marker\hat{D_i}^{clean}\to\hat{D_i}^{markdown}=(T,I,\tau)
\end{equation}
    
In image extraction, Marker detects images, diagrams, graphs and optimizes them, resulting in smaller files which facilitates the extraction of images from PDFs for subsequent steps by generating a markdown file including image file name and URLs. The full configurations for Marker are described in Appendix~\ref{apdx:marker}.

\textbf{Image and Table Processing\\}
One key improvement is generating captions to the images which alleviates image modality and turns the input into LLM’s native domain: text. This enhances the readability of markdown and improves the accuracy of the QA system by providing additional text information which can be indexed and retrieved. The image captioning in our project utilizes the LLaVA (Large Language and Vision Assistant) model, which is a fine-tuned version of the LLaMA/Vicuna model~ \cite{liu2023llava, liu2023improvedllava}. Every image in the markdown is given a unique ID (e.g.: image\_1.png) and the generated captions include these references, ensuring that images are identified and described. During markdown conversion, tables are represented in markdown syntax initially which can be verbose and inefficient. To enhance this, we compress markdown tables as dictionary format for efficient storage, reducing the storage space leading to easier data access, understanding and manipulation.

\subsubsection{\textbf{Embedding and storage}}

The first step involves breaking down the processed document $\hat{D_i}^{markdown}$ into smaller chunks $\{c_1,c_2,...c_m\}$. Data consisting of text, image captions and dictionary-formatted tables into 
separate segments. Each chunk $C_j$  is embedded into a high-dimensional vector space using an embedding model $f_{embed}$:

\begin{equation}
    f_{embed}(c_j)\to \mathit{v}_j \in \mathbb{R}^d
\end{equation}

These embeddings are then stored in a searchable database (Elasticsearch):

\begin{equation}
    Elasticsearch({\mathit{v}_1,\mathit{v}_2,...\mathit{v}_m})
\end{equation}

To enhance the retrieval process, we employ Recursive Abstractive Processing for Tree-Organized Retrieval (RAPTOR) for semantic indexing. It creates a tree structured index based on semantic content of the chunks for precise retrieval.

\subsubsection{\textbf{Question Answering}}

\textbf{RAG-Aware LLM Finetuning} 

To make the LLM adapt to the document domain as well as being aware of our Markdown format, image, and table structure, we trained RAG-Llama3-70B using a systematic process. We applied the same preprocessing steps of our system for training PDF documents, and then split the text into chunks of 5000 characters each. Using GPT-4, we generated relevant and comprehensive 
questions for each context chunk. These questions, along with their respective context, were fed back into GPT-4 to generate detailed answers, resulting to approximately 2000 question-answer pairs. To reflect the inference scenario where 10 chunks of 1000 characters were retrieved and 
appended to the prompt, we split the original context into five 1000-character chunks, mixed randomly with additional five 1000-character chunks from other documents, and appended as context to the question. The prompt details can be found in Appendix~\ref{apdx:A}. This also helps enhance robustness of the model on assessing relevance of the retrieved context. We finetune Llama3-70B-Instruct~\cite{grattafiori2024llama3herdmodels} with Low-Rank Adaptation (LoRA) \cite{a3} over 2 epochs, batch size of 8 and learning rate of 0.00008. All training hyperparameters can be found in Appendix~\ref{apdx:B}. The training took 8 hours on a single A100 80GB.

\textbf{Query Processing and Retrieval}

Upon receiving a query $q$, the query is also embedded into the same high-dimensional vector space:

\begin{equation}
    f_{embed}(q) \to \mathit{v}_q \in \mathbb{R}^d
\end{equation}

The system performs a similarity search to retrieve the top-k relevant chunks
${c_{j1},c_{j2},...c_{jk}}$ based on their cosine similarity to $\mathit{v}_q$

\begin{equation}
    Retrieve(\mathit{v}_q,\{\mathit{v}_1,\mathit{v}_2,...\mathit{v}_m\})\to \{c_{j1},c_{j2},...c_{jk}\}
\end{equation}

\textbf{Prompt Injection and Answer Generation}

The retrieved chunks are appended to the query in the prompt. The LLM then processes the prompt which now includes the user query and the injected chunks, to generate a response. The retrieved chunks $\{c_{j1},c_{j2},...c_{jk}\}$ are used as context to generate an answer $a$ using an LLM. The LLM is prompted with the query $q$ and the retrieved chunks:

\begin{equation}
    LLM(q,\{c_{j1},c_{j2},...c_{jk}\}) \to a
\end{equation}

We prompt the model to include the image ID related to the answer from the retrieved context in a specific format, i.e., [image\_1.png] for the later image recovery process. If the LLM answer contains image reference IDs, we retrieve the corresponding image from the database and display it to the user. In addition to handling images, if the LLM’s output includes dictionary-formatted tables, we extract and convert them back to markdown format for user display.

\section{{Experiments}}
\label{sec:Experiments}
\subsection{\textbf{Data Collection}}
Our dataset consists of 8 private internal documents from a production company. We used 6 of them for finetuning our model as described in Section 3, leaving the rest for testing. From the test documents, we constructed a test set by manually prompting GPT-4o to generate questions and answers based on some specific contexts, resulting in 100 question-answer pairs. The question covers a wide range of topics and formats involving text, images, tables to assess the system’s performance effectively.

\subsection{\textbf{Metrics}}
We employed several metrics to measure the effectiveness of our system. \textbf{Similarity} between the generated answers and the gold standard answers was assessed using embeddings from the GTE-large model. Additionally, we used accuracy at different thresholds - \textbf{accuracy@0.85}, \textbf{accuracy@0.9} and \textbf{accuracy@0.95} to evaluate the precision of the system. An answer is considered correct if its similarity score exceeds the given threshold score. These metrics provide a nuanced view of the system’s performance.

\section{{Results}}
\label{sec:Results}
\subsection{\textbf{Comparison with Baseline}}
To benchmark our system, we compared it with a baseline system that employs a similar approach. The baseline system parses the PDF files into plain text and saves these text chunks into a database. The retrieval process involves fetching these chunks and including them in the LLM prompt using either GPT-3.5-turbo or GPT-4o. Unlike our system, the baseline does not incorporate preprocessing steps such as header/footer removal or markdown conversion, nor does it retrieve images, diagrams and tables. Both systems were evaluated using the same set of 100 test questions. We measured the performance using the similarity metric and the accuracy at different thresholds (accuracy@0.85, accuracy@0.9 and accuracy@0.95).

\begin{table}[H]
\footnotesize
\setlength{\tabcolsep}{2.6pt}
\renewcommand{\arraystretch}{1.5}
\centering
\begin{tabular}{|p{1.3cm}|p{2.3cm}|p{1.4cm}|p{1.3cm}|p{1.3cm}|p{1.3cm}|}
\hline
\textbf{System} & \textbf{LLM Agent} & \textbf{Similarity} & \textbf{Accuracy @0.85} & \textbf{Accuracy @0.9} & \textbf{Accuracy @0.95} \\
\hline
Baseline & GPT-3.5-turbo & 0.8639 & 0.5889 & 0.3667 & 0.060 \\
\hline
Baseline & GPT-4o & 0.8647 & 0.6444 & 0.4000 & 0.1111 \\
\hline
PIER-QA & GPT-3.5-turbo & 0.8666 & \textbf{0.7640} & 0.3708 & 0.1124 \\
\hline
PIER-QA & GPT-4o & \textbf{0.8837} & 0.7191 & \textbf{0.4944} & \textbf{0.191} \\
\hline
\end{tabular}
\caption{Scores comparison with Baseline}
\label{tab:Table_1}
\end{table}

The results in Table \ref{tab:Table_1} demonstrate a clear performance advantage of the PIER-QA system over the baseline. Notably, PIER-QA achieved higher similarity scores and accuracy across all thresholds (0.85, 0.9, and 0.95) with both GPT-3.5-turbo and GPT-4o agents. This improvement is attributed to the enhanced preprocessing steps, including header/footer removal and markdown conversion, as well as the effective retrieval and integration of images, diagrams, and tables. These advancements enabled PIER-QA to generate more accurate and relevant responses, particularly at higher accuracy thresholds, highlighting its capability to handle complex PDF-based questions comprehensively.

\subsection{\textbf{Investigation of different LLM agents}}
To understand the impact of different language models on our system's performance, we evaluated the system using three different LLM agents: GPT-4o, GPT-3.5-turbo and Llama3-70B-Instruct, and our RAG-Llama3-70B. Each LLM agent was integrated into our system and the same set of 100 questions are used for evaluation. It's also worth noting that the Llama3-70B-Instruct and our RAG-Llama3-70B were quantized at 4-bit precision for efficiency.

\begin{table}[h]
\footnotesize
\setlength{\tabcolsep}{2.6pt}
\renewcommand{\arraystretch}{1.3}
\centering
\begin{tabular}{|p{3.3cm }|p{1.5cm }|p{1.5cm }|p{1.5cm }|p{1.5cm }|}
\hline
\textbf{LLM Agent } & \textbf{Similarity} & \textbf{Accuracy @0.85} & \textbf{Accuracy @0.9} & \textbf{Accuracy @0.95} \\
\hline
GPT-3.5-turbo  & 0.8666 & \textbf{0.7640} & 0.3708 & 0.1124 \\
\hline
GPT-4o & \underline{0.8837} & 0.7191 & \textbf{0.4944} & \textbf{0.1910} \\
\hline
Llama3-70B-Instruct & 0.8156 & 0.5280 & 0.2921 & 0.1348 \\
\hline
RAG-Llama3-70B & \textbf{0.8771} & \underline{0.7303} & \underline{0.4719} & \underline{0.2135} \\
\hline
\end{tabular}
\caption{Scores comparison of different LLM agents. Best scores are highlighted in \textbf{bold} while second best scores are \underline{underlined}.}
\label{tab:Table_2}
\end{table}

In Table \ref{tab:Table_2}, GPT-4o achieved the highest similarity score of 0.8837 and the best accuracy at 0.9, highlighting its strong retrieval and question-answering capabilities. However, our RAG-Llama3-70B model demonstrated notable performance as well. It outperformed the other models in terms of accuracy at the highest threshold – 0.9, and achieved the second-best scores on the other metrics, closely trailing GPT-4o even at 4-bit precision. This underscores the effectiveness of our RAGAware finetuning in adapting the model into the document domain and our system structure.

\subsection{\textbf{Table/Image Retrieval Performance}}
Given the importance of accurately retrieving and presenting nontextual information, we conducted experiments specifically focused on the performance of image and table retrieval. For this, we assessed the capability of the system to correctly identify and include tables and images in the generated answers.

We created 50 questions from the two test documents asking about specific images, and another 50 questions asking about specific tables. Image and table accuracy were measured by detecting whether the output of the system contains the correct image ID and table ID, respectively. We used our finetuned RAG-Llama3-70B model as the LLM agent in this experiment.

In Table \ref{tab:Table_3}, our PIER-QA system achieved considerable accuracies for image and table retrieval, at 65.66\% and 48.38\%, respectively. The results demonstrate the reliability of our approach in handling complex document structures with not only text but also tables and images without the use of multimodal LLMs.

\begin{table}[h]
\footnotesize
\setlength{\tabcolsep}{5pt}
\renewcommand{\arraystretch}{1.5}
\centering
\begin{tabular}{|p{2.0cm}|p{2.0cm}|}
\hline
\textbf{Task} & \textbf{Accuracy} \\
\hline
Image retrieval & 0.6566 \\
\hline
Table retrieval & 0.4838 \\
\hline
\end{tabular}
\caption{Table/Image Retrieval Performance}
\label{tab:Table_3}
\end{table}

\section{Limitations}
DBSCAN clustering algorithm might not always differentiate headers, footers and the main content, especially in highly variable PDF document formats. This misidentification could lead to loss of essential information in the further processing. The Markdown's tools performance in the conversion of complex PDF structures might not always be perfect, especially for complex tables, which leads to format errors or information loss. Additionally, our system always retrieve for every query, leading to potentially unnecessary retrieval steps and latency overhead. Research work shall be incorporated in the future to overcome the mentioned limitations with advanced tools and technologies. 

\section{Conclusion}
\label{sec:Conclusion}
In conclusion, our work significantly extends the powers of RAG frameworks by dealing with different data modalities in PDFs, such as text, images, vector diagrams, graphs, and tables. We have shown that our enriched RAG-based QA system could handle complex, multifaceted queries at higher accuracies and relevancies in incorporating and handling non-textual data. We further improved the system by finetuning the open-source LLM, closing the gap with the state-of-the-art close-source GPT-4o. Finally, we discussed some limitations of our system, which is a basis for future improvements.

\bibliography{neurips_2024_main_ref}
\bibliographystyle{plainnat}

\appendix
\section{Training data generation prompts}
\label{apdx:A}
\begin{figure}[H]
    \centering
    \includegraphics[width=0.5\textwidth]{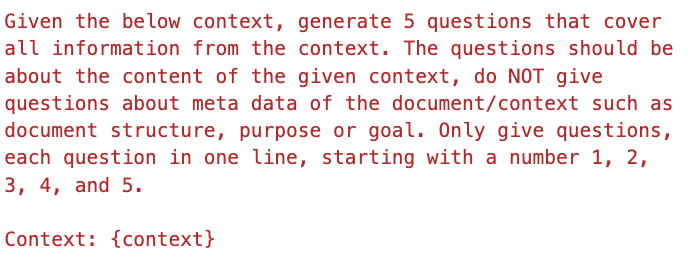} 
    \caption{Question generation prompt}
    \label{fig:gen_prompt}
\end{figure}

The data generation consists of two steps: question generation and answer generation. We split the preprocessed training documents into chunks (note that these chunks are longer than the retrieved chunks during inference) and put them into the `context' slot of the question generation prompt as shown in Figure~\ref{fig:gen_prompt}.

\begin{figure}[H]
    \centering
    \includegraphics[width=0.5\textwidth]{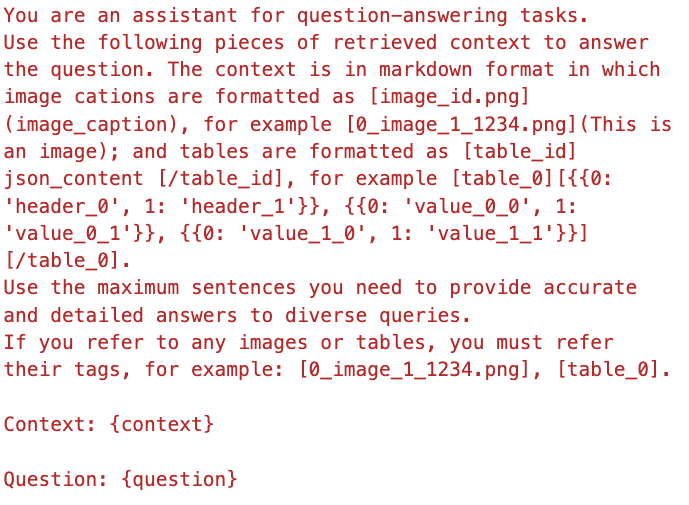} 
    \caption{Answer generation prompt}
    \label{fig:ans_gen_prompt}
\end{figure}

In the next step, we put generated questions and their corresponding context into answer generation template (Figure~\ref{fig:ans_gen_prompt}), resulting to question-answer pairs for the given context. The instruction in this template is also used for training and inference.

\section{DBSCAN hyperparameter details}
\label{apdx:dbscan}
DBSCAN hyperparameter values were selected based on empirical analysis to balance the precision and recall of header/footer removal.

\begin{itemize}
    \item $min\_samples$: This parameter represents the minimum number of samples in a neighborhood for a point to be considered as a core point. The value is dynamically set based on the number of pages in the PDF document:
    \begin{itemize}
        \item For documents with 6 pages or fewer: $min\_samples = 2$
        \item For documents with 7 to 8 pages: $min\_samples = 3$
        \item For documents with more than 8 pages: $min\_samples = 4$
    \end{itemize}
    This dynamic adjustment helps in better identification of headers and footers across documents of varying lengths.
    \item $eps$ (epsilon): This parameter defines the maximum distance between two samples for one to be considered as in the neighborhood of the other. A smaller value of `0.01` is chosen to ensure that only closely located text blocks (typically headers and footers) are clustered together.
\end{itemize}

\section{Marker configurations}
\label{apdx:marker}
The Marker tool in our PDF processing pipeline is responsible for extracting and handling various types of content from PDFs, including text, images, and layouts. By default, it uses the `Surya' OCR engine for efficient and accurate text recognition. The tool is configured to run OCR on all pages of a PDF, even if some text can be directly extracted, ensuring comprehensive text recognition across the entire document. 

In addition to text recognition, the Marker tool employs advanced models for layout detection and text ordering, such as the Texify model. A post-processing model further refines the data by applying a probability threshold to ensure only high-confidence predictions are retained, reducing errors and enhancing the overall quality of the output. The full settings for Marker can be found in \ref{tab:marker}.

\begin{table*}[]
\centering
\resizebox{\textwidth}{!}{%
\begin{tabular}{|l|l|l|}
\hline
\textbf{Parameter} &
  \textbf{Default} &
  \textbf{Description} \\ \hline
IMAGE\_DPI &
  96 &
  DPI for rendering images from PDFs. \\ \hline
EXTRACT\_IMAGES &
  TRUE &
  Whether to extract images from PDFs. \\ \hline
PAGINATE\_OUTPUT &
  FALSE &
  Whether to paginate the output markdown. \\ \hline
DEFAULT\_LANG &
  English &
  \begin{tabular}[c]{@{}l@{}}Default language for OCR, should match a key\\ in TESSERACT\_LANGUAGES.\end{tabular} \\ \hline
DETECTOR\_BATCH\_SIZE &
  None &
  \begin{tabular}[c]{@{}l@{}}Batch size for text line detection. Defaults to 6 \\ for CPU, 12 otherwise.\end{tabular} \\ \hline
SURYA\_DETECTOR\_DPI &
  96 &
  DPI for the Surya detector. \\ \hline
INVALID\_CHARS &
  {[}chr(0xfffd){]} &
  Characters to ignore during OCR. \\ \hline
OCR\_ENGINE &
  "surya" &
  \begin{tabular}[c]{@{}l@{}}OCR engine to use, defaults to "surya" on GPU \\ and "ocrmypdf" on CPU.\end{tabular} \\ \hline
OCR\_ALL\_PAGES &
  FALSE &
  \begin{tabular}[c]{@{}l@{}}Whether to run OCR on every page even if text \\ can be extracted.\end{tabular} \\ \hline
SURYA\_OCR\_DPI &
  96 &
  DPI for Surya OCR. \\ \hline
RECOGNITION\_BATCH\_SIZE &
  None &
  \begin{tabular}[c]{@{}l@{}}Batch size for Surya OCR, defaults to 64 for \\ CUDA, 32 otherwise.\end{tabular} \\ \hline
TESSERACT\_TIMEOUT &
  20 &
  Timeout for Tesseract OCR. \\ \hline
TEXIFY\_MODEL\_MAX &
  384 &
  Max inference length for Texify. \\ \hline
TEXIFY\_TOKEN\_BUFFER &
  256 &
  \begin{tabular}[c]{@{}l@{}}Number of tokens to buffer above max for \\ Texify.\end{tabular} \\ \hline
TEXIFY\_DPI &
  96 &
  DPI for rendering images in Texify. \\ \hline
TEXIFY\_BATCH\_SIZE &
  None &
  \begin{tabular}[c]{@{}l@{}}Batch size for Texify, defaults to 6 for CUDA, \\ 12 otherwise.\end{tabular} \\ \hline
TEXIFY\_MODEL\_NAME &
  "vikp/texify" &
  Name of the Texify model. \\ \hline
SURYA\_LAYOUT\_DPI &
  96 &
  DPI for Surya layout. \\ \hline
BAD\_SPAN\_TYPES &
  \begin{tabular}[c]{@{}l@{}}{[}"Caption", "Footnote",\\ "Page-footer", "Page-header",\\  "Picture"{]}\end{tabular} &
  Types of spans to consider as bad spans. \\ \hline
LAYOUT\_MODEL\_CHECKPOINT &
  "vikp/surya\_layout3" &
  Checkpoint for the layout model. \\ \hline
BBOX\_INTERSECTION\_THRESH &
  0.7 &
  Threshold for bounding box intersection. \\ \hline
LAYOUT\_BATCH\_SIZE &
  None &
  \begin{tabular}[c]{@{}l@{}}Batch size for layout model, defaults to 12 for \\ CUDA, 6 otherwise.\end{tabular} \\ \hline
SURYA\_ORDER\_DPI &
  96 &
  DPI for Surya ordering. \\ \hline
ORDER\_BATCH\_SIZE &
  None &
  \begin{tabular}[c]{@{}l@{}}Batch size for ordering model, defaults to 12 \\ for CUDA, 6 otherwise.\end{tabular} \\ \hline
ORDER\_MAX\_BBOXES &
  255 &
  Maximum number of bounding boxes for ordering. \\ \hline
EDITOR\_BATCH\_SIZE &
  None &
  \begin{tabular}[c]{@{}l@{}}Batch size for final editing model, defaults to \\ 6 for CUDA, 12 otherwise.\end{tabular} \\ \hline
EDITOR\_MAX\_LENGTH &
  1024 &
  Maximum length for the final editing model. \\ \hline
EDITOR\_MODEL\_NAME &
  "vikp/pdf\_postprocessor\_t5" &
  Name of the final editing model. \\ \hline
ENABLE\_EDITOR\_MODEL &
  FALSE &
  Whether to enable the final editing model. \\ \hline
EDITOR\_CUTOFF\_THRESH &
  0.9 &
  \begin{tabular}[c]{@{}l@{}}Probability threshold to ignore predictions below \\ this value.\end{tabular} \\ \hline
\end{tabular}%
}
\caption{All Marker configurations}
\label{tab:marker}
\end{table*}

\section{Llama3 training hyperparameters}
We provide the full list of training hyperparameters for our RAG-Llama3-70B in Table~\ref{tab:params}.
\label{apdx:B}

\begin{table*}[]
\centering
\resizebox{\textwidth}{!}{%
\begin{tabular}{|l|c|l|}
\hline
\textbf{Parameter}      & \textbf{Value} & \textbf{Description}                                          \\ \hline
load\_in\_8bit          & FALSE          & Indicates whether to load the model in 8-bit precision.       \\ \hline
load\_in\_4bit          & TRUE           & Indicates whether to load the model in 4-bit precision.       \\ \hline
adapter                 & qlora          & Specifies the adapter type to use, in this case, QLoRA.       \\ \hline
sequence\_len           & 6000           & Maximum sequence length for training.                         \\ \hline
sample\_packing &
  TRUE &
  \begin{tabular}[c]{@{}l@{}}Enables efficient multi-packing with block diagonal \\ attention and per sequence position\_ids.\end{tabular} \\ \hline
pad\_to\_sequence\_len &
  TRUE &
  \begin{tabular}[c]{@{}l@{}}Pads inputs to ensure each step uses constant-sized \\ buffers, reducing memory fragmentation.\end{tabular} \\ \hline
lora\_r                 & 8              & Rank of the low-rank adaptation matrices in LoRA.             \\ \hline
lora\_alpha             & 16             & Scaling factor for LoRA.                                      \\ \hline
lora\_dropout           & 0.05           & Dropout rate for LoRA layers.                                 \\ \hline
lora\_target\_linear    & TRUE           & Indicates whether to target all linear modules in LoRA.       \\ \hline
gradient\_accumulation\_steps &
  4 &
  \begin{tabular}[c]{@{}l@{}}Number of steps to accumulate gradients before \\ updating model weights.\end{tabular} \\ \hline
micro\_batch\_size      & 2              & Number of samples in each micro-batch.                        \\ \hline
num\_epochs             & 2              & Number of epochs to train the model.                          \\ \hline
optimizer &
  adamw\_bnb\_8bit &
  \begin{tabular}[c]{@{}l@{}}Optimizer used for training, in this case, AdamW \\ with 8-bit precision.\end{tabular} \\ \hline
lr\_scheduler           & cosine         & Learning rate scheduler type, in this case, cosine annealing. \\ \hline
learning\_rate          & 8.00E-06       & Learning rate for training.                                   \\ \hline
train\_on\_inputs &
  FALSE &
  \begin{tabular}[c]{@{}l@{}}Indicates whether to include the human's prompt in \\ the training labels.\end{tabular} \\ \hline
group\_by\_length       & FALSE          & Whether to group data by length to minimize padding.          \\ \hline
bf16                    & auto           & Use bf16 precision if available.                              \\ \hline
gradient\_checkpointing & TRUE           & Enables gradient checkpointing to save memory.                \\ \hline
logging\_steps          & 20             & Frequency of logging training progress.                       \\ \hline
flash\_attention        & TRUE           & Enables flash attention for improved performance.             \\ \hline
warmup\_steps           & 20             & Number of steps for learning rate warmup.                     \\ \hline
evals\_per\_epoch       & 5              & Number of evaluations to perform per epoch.                   \\ \hline
saves\_per\_epoch       & 3              & Number of times to save checkpoints per epoch.                \\ \hline
\end{tabular}%
}
\caption{All training hyperparameters.}
\label{tab:params}
\end{table*}

\end{document}